# A Semantic Model for Historical Manuscripts

Sahar Aljalbout, Gilles Falquet

Centre Universitaire d'informatique, University of Geneva, Switzerland
{saharjalbout@gmail.com, Gilles.Falquet@unige.ch}

**Abstract.** The study and publication of historical scientific manuscripts are complex tasks that involve, among others, the explicit representation of the text meanings and reasoning on temporal entities. In this paper we present the first results of an interdisciplinary project dedicated to the study of Saussure's manuscripts. These results aim to fulfill requirements elaborated with Saussurean humanists. They comprise a model for the representation of time-varying statements and time-varying domain knowledge (in particular terminologies) as well as implementation techniques for the semantic indexing of manuscripts and for temporal reasoning on knowledge extracted from the manuscripts.

## 1 Introduction

The field of digital humanities is a collaborative and transdisciplinary area between digital technologies and the discipline of humanities [1]. This field is undergoing major changes in its methodology mainly because of the growing interest of scholars in high quality digital resources. In this paper, we present an advanced model to help humanists deal with the knowledge intensive tasks they must perform when studying historical scientific manuscripts. For this purpose we propose a set of evolving interconnected knowledge resources (ontologies, terminologies, transcriptions etc.) that represent the current state of our knowledge about a corpus of manuscripts. The model was applied on the manuscripts of Ferdinand de Saussure.

Ferdinand de Saussure (1857-1913) is a Swiss linguist [2] and semiotician. He has been known for his work in general linguistics as well as for his contributions in the rather more exclusive field of comparative grammar. Despite his excellent reputation in the field of linguistic, he published very little. For instance, he never published the theory he developed in the *course of general linguistics* he taught three times and which is considered as the work of his life. It is on the basis of lecture notes of his students that the book *Course in General Linguistics* (Cours de linguistique générale CLG) was published in 1916. The legacy of Saussure is fortunately not limited to these monographs but includes a fund of about 50,000 handwritten pages deposited in the libraries of Geneva, Paris and Harvard. Recently, all these pages were photographed using a high definition digital camera to be used in the project. These manuscripts are of a major importance for the Saussureans (scholars working on Saussure's manuscripts). They constitute the only way to reach a better understanding of Saussure's ideas.

The major challenge for Saussureans is to understand the content of these manuscripts. The difficulty of the manuscripts handling is a key issue. As of today, only 5,000 manuscripts of the 50,000 pages have been transcribed, and published in several publications [3]. In order to be able to process the text contained in the manuscripts, we work directly on the transcriptions while keeping the direct reference to the corresponding manuscripts. A study has been conducted with three internationally recognized Saussurean experts, who are members of the *Cercle Ferdinand de Saussure*[1]. The aim was to study the current methodology adopted by Saussureans to process manuscripts in order to discover their needs and to get familiar with the problems they face. Several needs were detected. The main problem with these manuscripts relates to the fact that they have been produced in different periods with several views and perspectives. They are sometimes incomplete or not well organized, what makes the meaning of data unclear without interpretations. Furthermore, due to drifting concepts in time, the context can be subject to different interpretations. In order to deal with the vast amount of data, a classification of the manuscripts into meaningful thematic categories would be required. In addition, one of the most important aims of the Saussureans is to tag the manuscripts with their writing time and space. This enrichment can help the historical or epistemological research work such as the establishment of a clear idea of Saussure's work. It can be done by following the bibliographic references, names, events, and institutions that Saussure used to quote in his scriptures.

The remainder of this paper will be divided as follow. In section 2, we introduce the scholars' requirements in terms of structures and operations. Section 3 reviews related works in the field of digital humanities. The proposed model is presented in section 4 and the implementation issues are discussed in section 5. Finally, we conclude the work in section 6.

## 2      Requirements

The analysis of the Saussurean's requirements showed that they can be organized along four main dimensions:

*Retrieving and accessing*. The scholars wants to look through the sources according to specific thematic categories. In this respect, the classification schemas designed by the archivists is clearly insufficient. Several search criteria have to be taken into consideration. Tools for the visualization of the manuscripts are also required.

*Understanding*. The terminology, in particular the definitions used by Saussure vary over time, making some texts hard to understand or leading to misunderstanding. This is indeed a very common fact in scientific works that span a long period of time and that explore (and define) new scientific areas. Furthermore, manuscripts contains terminological knowledge with a dynamic temporal perspective that makes them difficult to understand. A recent work [11] shows the existence of 16 terminologies used by Saussure. The terminology is time bound, which makes formalization difficult and results dependent on different interpretations.

---

[1] Two of them are members of the editorial board of the international journal "Cahiers Ferdinand de Saussure".

*Dating*. The chronological order of the manuscripts is of paramount importance to understand the author's work. The challenges here are on how to infer the dates from the existing entities.

*Publishing*. Disclosing the author's work by means of a scholarly publication of a set of manuscripts is also an important target.

Refining this analysis then lead to structural and functional requirements.

**Structural requirements: a Multi-Knowledge Resource Model**
The manuscript knowledge base must contain representations of the manuscripts, their transcriptions, and domain knowledge related to the manuscript content. A key issue for the scholars when dealing with Saussure's manuscripts is to precisely date them or at least to rebuild their writing sequence. Three basic types of indicators can help solving this problem: 1) references to persons, places, and events; 2) the terminology used by the author; 3) the bibliographic references to other works. Therefore the knowledge base (Fig.1) should contain:
− *Representation of the manuscript contents*
− *Contextual knowledge that describe persons, places, events and other entities.*
− *Terminological knowledge that describe the meaning the author associated with terms at a given point in time;*
− *Terminological knowledge that describe the concepts of a scientific (sub)domain*
− *Bibliographic knowledge about the works cited in the manuscripts.*

**Functional Requirements**
These requirements are divided into four main functionalities.

*Creating ontologies as a mean to address the lack of formalization.* The first problem we encountered when modeling historical information was the lack of ontologies to describe some historical facts. Although some ontologies like CIDOC-CRM[2] (i.e. to describe concepts and relationships in cultural heritage) and *Simple Event Model*[3] (i.e. to describe events) etc. were developed, they are insufficient to: (a) describe the different time periods during which an entity holds and the reason that initiates and terminates these different periods (b) reason on time or to compare time entities (c) model the interpretations and evidences required for historians (d) represent the temporal terminology series.

*Data integration using semantic linking operations.* Linking datasets by mean of different type of semantic links enrich the information contained in them contributing to the goal of linked data. Different type of links should exist between documents and contextual/terminological entities. The requirement concerning these links is to provide automated or semi-automated tools to generate them.

*Inferencing of implicit knowledge using ontologies and documents.* Historical ontologies are used to control the data models in which the data are represented. But from the Semantic Web perspective, one of the main promises of these ontologies is to explicit implicit information by providing a way to express inference rules, in particular time related inferences.

---

[2] http://www.cidoc-crm.org/Version/version-6.2
[3] http://semanticweb.cs.vu.nl/2009/11/sem/semdoc.html

*Information Retrieval and visualization.* Sophisticated information needs semantically enhanced search operations that take advantage of the contextual and background knowledge associated to the manuscripts. These query based search operations must be complemented with navigation-based search that utilizes the semantic links between the resources.

## 3    Related Work

Many systems are generally equipped with a web interface to access digital manuscript collections. The "Nietzsche Source[4]" project is dedicated to the scholarly publication of F. Nietzsche's works. It may be noted however that there is no link between the critical editions and digital reproductions. Furthermore, it features an efficient search engine on passages. We can find similar features in the PhiloLogic[5] project which is used as the primary full text search and retrieval tool developed by the ARTFL[6] project .The aim of the Bentham Project[7] is to transcribe the manuscripts of the philosopher and jurist Jeremy Bentham using crowdsourcing. Transcripts are encoded with TEI[8] and the entire interface of the site is based on MediaWiki[9] with a modified skin.

On the other hand, the ontological approach has been adopted to improve the access quality to the manuscript content. They used such techniques in the Sharing Ancient Widsoms (SAWS) project in which semantic information is extracted from TEI documents [4]. These documents are the fundamental components of a conceptual network searchable by researchers. The general collection theme is about the ancient Greek and Arabic wisdom. Other projects [5] have semi-automatically built ontologies that gather knowledge about manuscripts. Some projects [6] worked on providing researchers with innovative ways for accessing and sharing digital manuscripts.   Knowledge representation techniques can also be used in particular formal or semi-formal ontologies to index, classify, represent the semantics of documents, or help performing knowledge intensive tasks. For instance, [7] describe an ontology-based digital library that allows researchers to describe and debate the contribution a document makes and its relationship to the literature.

Some of the above mentioned projects and others rely on TEI (Text Encoding Initiative) which is a consortium whose purpose is to propose recommendations for encoding digital documents [8]. The TEI document type definition provides a very precise description of many types of documents: general texts; poetry; speech transcription; dictionaries; manuscripts; tables; graphs; networks.

---

[4] http://www.nietzschesource.org/
[5] https://sites.google.com/site/philologic3/home
[6] http://artfl-project.uchicago.edu/
[7] http://www.ucl.ac.uk/Bentham-Project
[8] http://www.tei-c.org/index.xml
[9] https://www.mediawiki.org/wiki/MediaWiki

An interesting review [12] covers current techniques on how to apply semantic technologies to historical research. The analysis of the historical field is divided into twofold: the identification of the problems and the presentation of several relevant projects. All the projects presented in the review show at some point that the semantic web technologies can solve various historical problem such as: knowledge formalization and enrichment, data integration, mapping and flexibility etc... Furthermore, a recent project [13] provide metadata and links to directly access the digital content of various cultural heritage across Europe. The goals of the project are: (i) to transform metadata and content of cultural heritage into Europeana data model (EDM) and (ii) the creation of tools and services to reuse the data in digital humanities.

## 4 A Semantic Model for Historical Scientific Manuscripts

In this section, we present a semantic model for a corpus of manuscripts. This model contains a set of evolving interconnected knowledge resources that represent the current state of our knowledge about the corpus.

A UML class diagram of the proposed model is shown in Fig.1. One of our objectives is to publish knowledge about the manuscripts as linked open data. The prerequisite of that is to formalize the model as OWL .The components of the model are the following.

### 4.1 Documents and Knowledge Resources

*Manuscripts.* The manuscripts are the center of the model. A manuscript is composed of writing surfaces (e.g. pages of a notebook, sides of a note card, etc.). Librarians assign each manuscript with a unique ID. This ID is used to create the URI of the manuscript. The surfaces may also be explicitly identified by a number that is unique within the manuscript.

*Transcriptions.* We consider a transcription as the as exact as possible copies of the manuscripts, reproducing its content letter by letter. It can also reproduce the layout of the manuscript showing the subsequently made corrections. For each manuscripts we can have several transcriptions. An instance of the class Transcription represents the transcription of a single zone within a surface that belongs to a manuscript. The transcription of all parts of the manuscript is a text made of a sequence of transcriptions. In many instances, this sequence does not correspond to the physical sequence of the surfaces or zones, it is a reconstruction of the intended text flow of the author.

*Scientific Publications.* They refer to published or unpublished books, articles, or other documents that have strong connection with the manuscripts. A scientific document can be written by Saussure or referring to Saussure's work.

*Annotations.* Annotations are comments or clarifications made by the transcribers in order to explain better or comment a part of the transcription.

*Knowledge resources.* A novel aspect of this infrastructure is its ability to store and interconnect different types of knowledge resources. Some might recommend the use of a single representation language, such as the OWL ontology language, to represent all the knowledge resource types. Nevertheless, this approach would reduce the

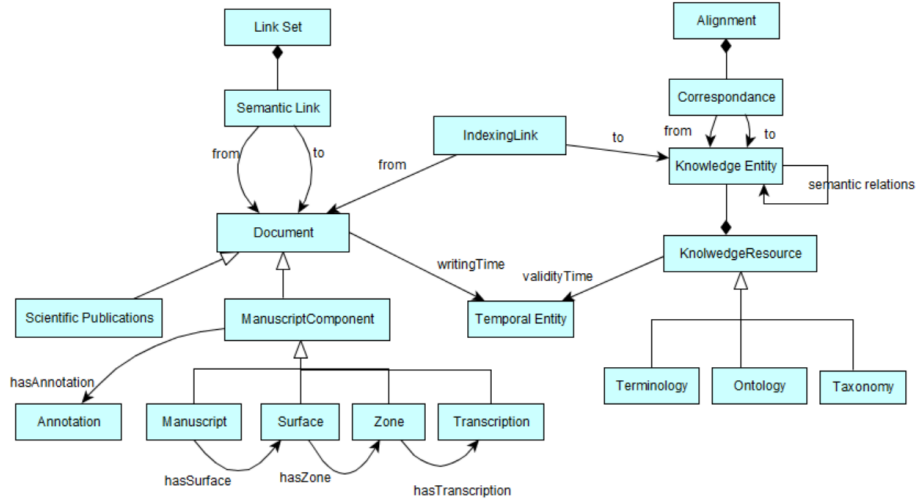

*Figure 1 Infrastructure Class Diagram*

usability and interest of these resources. In fact, as argued in [9], the diversity of knowledge representation must be preserved because each one of them has specific characteristics that cannot be represented with another one. For instance, the *broader-term* hierarchical relations in a thesaurus may express a generic/specific relation, or a part/whole relation, or a class/instance relation. Similarly, for terminologies expressed in the standard TBX format there is no canonical mapping to OWL (e.g. concepts can be mapped to OWL classes with annotation properties or to individuals with object properties). Therefore, we consider an abstract view of all these formalisms by considering the modeling primitives of each formalism as subclasses of the *KnowledgeEntity* class or as subproperties of the *KnowledgeEntityRelation* property. For example:

*For SKOS thesauri (sism: is the namespace for the Semantic infrastructure for scientific manuscripts vocabulary):*

skos:ConceptScheme **rdfs:subClassOf** sism:KnowledgeResource .
skos:Concept **rdfs:subClassOf** sism:KnowledgeEntity .
skos:inScheme **rdfs:subPropertyOf** sism:inKnowledgeResource .
skos:Scheme **rdfs:subClassOf** sism:KnowledgeResource .
skos:semanticRelation **rdfs:subPropertyOf** sism:semanticRelation

*For OWL ontologies:*

owl:Ontology **rdfs:subClassOf** sism:KnowledgeResource .
owl:NamedIndividual **rdfs:subClassOf** sism:KnowledgeEntity .
owl:Class **rdfs:subClassOf** sism:KnowledgeEntity .
owl:ObjectProperty **rdfs:subClassOf** sism:KnowledgeEntity .
owl:DatatypeProperty **rdfs:subClassOf** sism:KnowledgeEntity .

rdfs:subClassOf **rdfs:subPropertyOf** sism: semanticRelation.

The co-existence of multiple conceptualizations in the same knowledge base raises the question of possible inconsistencis. What happens if two resources contain contradictory assertions about the same concept? The rather circumspect approach [9] we propose here is to consider that the concepts defined in different resources are all different. Nevertheless, entities in different resources may be interconnected through alignments made of correspondences of the form (*entity*$_1$, *entity*$_2$, *relation*, *confidence*) between entities. The alignments play at least two role: 1) the participate in the consistency checking when loading new resources and 2) the help the user navigate among the resources.

### 4.2 Temporal Entities and Properties

A critical problem for the representation of temporal knowledge in the semantic web domain is to deal with temporal properties whose value is likely to change. Thus we distinguish between static properties and the *fluents*[10]. We consider a fluent as a relation that takes place in an interval of time and not in others. For example, the statement "Saussure was born in 1857" is considered as a static statement, however the statement "Saussure lived in Germany between 1876 and 1881" is only true for a defined time interval (i.e. then false, then true again, etc.) and is considered as a fluent. In a semantic web perspective, the problem of fluent representation is related to representation languages such as OWL which only support binary relations. The idea of the fluent representation is to add an object *FluentRelation* between the subject and the object that will support the temporal properties. In general, an event can initiates a fluent and can also terminate it.

Figure 2 shows the various time-related entities of the model. For instance, documents in some cases have a writing date (that appears in the document) that is represented by the *writingTime* property. Otherwise, the date must be inferred using temporal reasoning. An inferred writing time is represented by the *inferredWritingTime* property.

Understanding the text implies recognizing the meaning of words. However, since the corpus spans a long period of time, the meaning of some terms can vary greatly. Therefore, we propose to use a fluent *uses* that associates a person (an author in this case) with a terminology (composed of a set of terms and their definitions). By using the writing period of a document, it is then possible to find the meaning of the terms (since the terminologies are dated) used and thus to index them semantically in the most correct way.

In addition to the terminologies, the contextual knowledge can typically be comprised of several ontologies such as:

– *people ontology : describe all the persons or characters cited in the manuscripts and the relations that can exist between them. Persons can be members of Saussure's family or scholars he cites in his writings*

---

[10] https://en.wikipedia.org/wiki/Event_calculus

- *literary people ontology: comprises all the literary personage whether religious, mythological or legendary that are cited in the manuscripts*
- *events ontology: describe the events related to Saussure's private life. An event can be personal, scientific or social event. It can also describe a series of events*

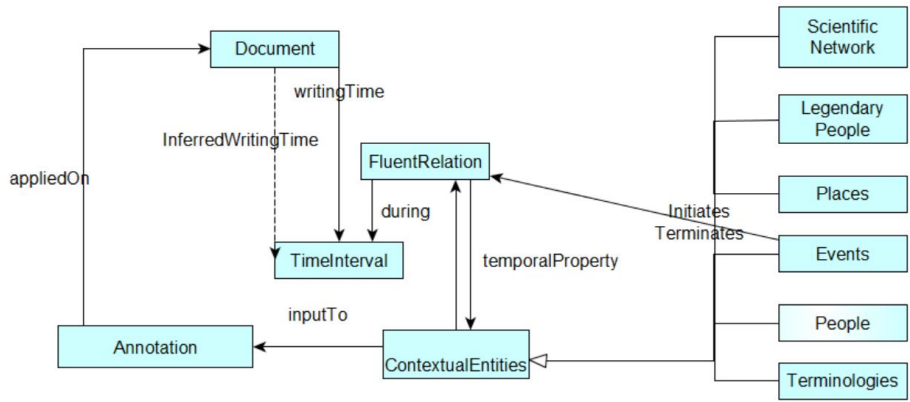

Figure 2 Enhanced Class Diagram

- *places ontology: describe all the places cited in the manuscript, ranging from streets to countries. These ontologies are typically derived from existing geographic ontologies like geonames[11]*
- *Scientific network ontology: describing the flow of bibliographic and citations entities that we can extract form the manuscripts*

## 5   Service Implementation

A set of services that correspond to the requirements identified with the Saussurean experts are provided on top of this infrastructure. The most important ones are:

*Importing knowledge resources.* The model does not impose the use of a particular representation language as mentioned in section 5.1. As a consequence, the importing procedure must handle different formal languages (OWL, SKOS, RDFS, etc.) representing different resource categories (ontologies, terminologies, thesauri, etc.) and generate the corresponding knowledge resource and knowledge entities in the knowledge base. The import operation proceeds as follows:

1. Create an RDF version of the resource file: assign semantically correct and non-conflicting URIs to the resource elements (e.g. generate a concept URI from the lexical form of a concept in a terminology); generate the necessary triples for defining the knowledge entities and for associating them with their knowledge resource (this associations is explicitly defined in SKOS thesauri, but not in OWL ontologies);
2. Import the RDF triples into the knowledge base
3. Check for inconsistencies. If the resource uses its own independent vocabulary no inconsistency problem can arise (unless the resource itself is inconsistent). Otherwise the resource, together with the already stored resources must be sent to a consistency checker (e.g. an OWL reasoner) to detect possible inconsistencies. In this case, the correspondences in the alignments are considered as equivalence and subsumption axioms that must be taken into account by the consistency checker.

---
[11] http://www.geonames.org/

Fig. 3 shows a sample navigation through terminologies, reaching two different terms corresponding to the word "phonation".

*Semantic indexing*. The semantic indexing of the manuscripts consists in connecting the words (or multi-word expressions) that appear in the transcriptions to the corresponding concepts in the knowledge resources (in particular in terminologies.)

As already mentioned, the difficulty of this task stems from the evolution of the terminology used by the author over his lifetime. Since the differences between two definitions of the same word in successive terminologies may be very subtle, it is extremely difficult to fully automate the semantic indexing process. Therefore, in the current implementation, the indexer computes and stores scores for the word - concept associations, keeping all the meanings whose score is higher than a threshold. These scores can then be shown in the user interfaces and used to sort the results or to help experts decide on the correct meanings.

The score of a word - concept association depends on two factors:

1. The similarity between the word's context in the transcription and the "contexts of use" provided for the concept in the terminology (see Fig.3). This corresponds to evaluating the similarity in terms of distributional semantics (in fact a very rough approximation of it)
2. If the writing time of a manuscript is known, the concepts that belong to terminologies used by the author at that time are given a higher score

*Temporal reasoning*. For the majority of the manuscripts, we do not know the date of writing which hinders the establishment of a clear and logical sequence of ideas. One of the promises of this project, in the Saussureans' perspective, is the enrichment of dataset with temporal knowledge. This enrichment can be done by using an inference system based on the bibliographic references, names, events… and institutions that Saussure used to quote in his scriptures. It is based on the generation of *FluentRelation* rules and purely temporal rules. Some inference rules can be directly expressed as SWRL rules, however, rules that generate fluent relations create new objects (new instances of *FluentRelation* and *time:Interval*). Therefore they cannot be expressed in SWRL. The adopted solution is to express this kind of rule as a SPARQL construct query that generates blank node. The general structure of such a query, for a property p is

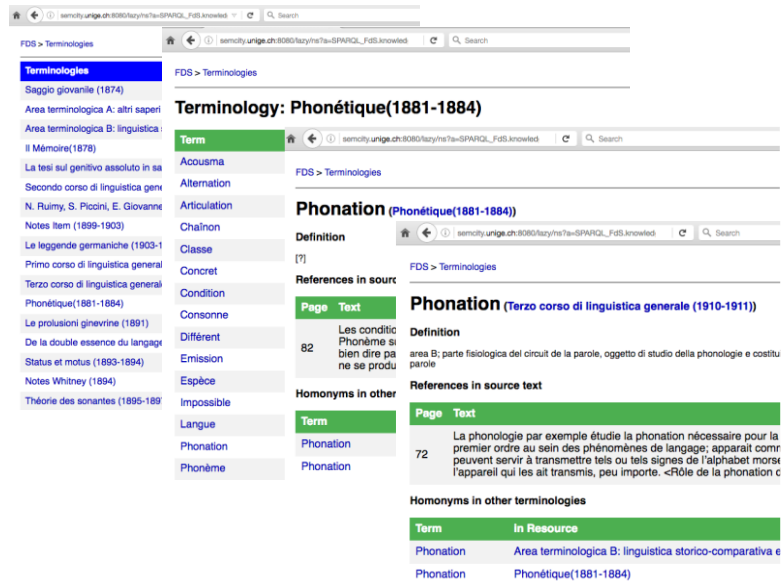

*Figure 3 Screenshot of the user interface*

```
construct {?X :p
            [a :FluentRelation; :p ?Y;
             :during [a time:Interval; ... ]]
          }
where { <<rule body>>
  filter not exists{ <<subsuming fluent>> }
}
```

A typical example is the rule "*if a manuscript M written by A is a letter to B and the writing time of M is [t1..t2] then A knows B during the interval [t1 .. end-of-considered-period]*". The corresponding query is

```
construct {?A :knows
      [a :FluentRelation ; :knows ?B ;
       :during [a time:Interval ; time:hasBeginning ?t1 ;
                time:hasEnd :end-of-considered-period]
      ]}
where { ?M a :Letter;  :author ?A; :to ?B;
           :writingTime [time:hasBeginning ?t1]
  filter not exists{
    ?t1 time:inXSDDate ?d1.
    :end-of-considered-period time:inXSDDate ?d2.
    ?A :knows
      [a :FluentRelation ; :knows ?B ;
```

```
        :during [time:hasBeginning/time:inXSDDate ?ed1;
                time:hasEnd/time:inXSDDate ?ed2]]
        filter(?ed1 <= ?d1 &&  ?d2 <= ?ed2)
    }
}
```

The filter not exists {...} part of the query guarantees that the generated fluent is not already subsumed by or equivalent to a fluent that represents the same relationship between the same individuals during a wider or equal time interval.

To ensure the completeness of the reasoning, once all the construct queries have been executed and the generated triples inserted into the knowledge base, the DL/SWRL reasoner is invoked, then the queries are re-executed, and so on until no new triple is generated.

The process is guaranteed to terminate because 1) SWRL rules do not generate new nodes and 2) there is only a finite number of dates in the considered time period and hence a finite number of time intervals, therefore the construct queries generate only a finite number of FluentRelation instances (because the queries never generate equivalent fluents).

## 6      Conclusion and future work

In this paper, we presented a model illustrating how semantic technologies can be applied to historical datasets. A study carried out with humanists led to the modeling of a temporal infrastructure for the storage and publication of a corpus of scientific manuscripts. This work allowed us to examine how semantic technologies can help in solving some of the problems the humanists face when studying these manuscripts. In particular, we proposed a multi-knowledge resource structure to represent the evolving nature of an author's terminology. We showed that a semi-automated semantic indexing technique can be employed for these "terminological time series". We also focused on the representation of time-varying properties and implemented a notion of fluent relation, directly inspired by the event calculus, and an appropriate inference mechanism that utilizes SPARQL queries.

We developed an experimental system that comprises an RDF knowlege base in which several hundred transcriptions and 15 terminological resources have been ingested. This system is coupled with the manuscript visualization and annotation system [10] that was developed in former project with the Saussurean experts.

In the near future we will 1) evaluate the usability and performance of the semi-automated semantic indexing technique and 2) work with the humanists to elicit temporal inference rules and to test them on a new, extended, version of the knowledge base. An interesting challenge here is to design a temporal rule language that is usable by the humanists.